\documentclass[10pt,twocolumn,letterpaper]{article}

\usepackage{iccv}
\usepackage{times}
\usepackage{epsfig}
\usepackage{graphicx}
\usepackage{amsmath}
\usepackage{amssymb}

\usepackage{subfigure}
\usepackage{caption}
\usepackage{multirow}
\usepackage{url}
\usepackage{algorithm}
\usepackage{algorithmic}
\usepackage{color}
\usepackage{booktabs}
\usepackage{bm}
\usepackage{dsfont}
\usepackage{float}
\usepackage{stfloats}
\usepackage{setspace}

\usepackage[pagebackref=true,breaklinks=true,letterpaper=true,colorlinks,bookmarks=false]{hyperref}

\iccvfinalcopy 

\def\iccvPaperID{959} 

\ificcvfinal\fi
\begin{document}

\title{Larger Norm More Transferable: An Adaptive Feature Norm Approach for Unsupervised Domain Adaptation}

\author{Ruijia Xu$^{1}$ \qquad Guanbin Li$^{1}$\thanks{Corresponding author is Guanbin Li.} \qquad Jihan Yang$^{1}$ \qquad Liang Lin$^{1,2}$\\
$^1$School of Data and Computer Science, Sun Yat-sen University, China
\\$^2$DarkMatter AI Research\\
xurj3@mail2.sysu.edu.cn \quad liguanbin@mail.sysu.edu.cn \quad jihanyang13@gmail.com \quad linliang@ieee.org}

\maketitle
\ificcvfinal\thispagestyle{empty}\fi

\begin{abstract}
Domain adaptation enables the learner to safely generalize into novel environments by mitigating domain shifts across distributions. Previous works may not effectively uncover the underlying reasons that would lead to the drastic model degradation on the target task. In this paper, we empirically reveal that the erratic discrimination of the target domain mainly stems from its much smaller feature norms with respect to that of the source domain. To this end, we propose a novel parameter-free Adaptive Feature Norm approach. We demonstrate that progressively adapting the feature norms of the two domains to a large range of values can result in significant transfer gains, implying that those task-specific features with larger norms are more transferable. Our method successfully unifies the computation of both standard and partial domain adaptation with more robustness against the negative transfer issue. Without bells and whistles but a few lines of code, our method substantially lifts the performance on the target task and exceeds state-of-the-arts by a large margin (11.5\% on \textit{Office-Home}~\cite{10_venkateswara2017deep} and 17.1\% on \textit{VisDA2017}~\cite{6_peng2017visda}). We hope our simple yet effective approach will 
shed some light on the future research of transfer learning. 
Code is available at~\url{https://github.com/jihanyang/AFN}.
\end{abstract}


\begin{figure}[t]
    \centering
    \setlength{\belowcaptionskip}{-0.5cm}  
    \setlength{\abovecaptionskip}{0.1cm}  
    \includegraphics[width=0.48\textwidth]{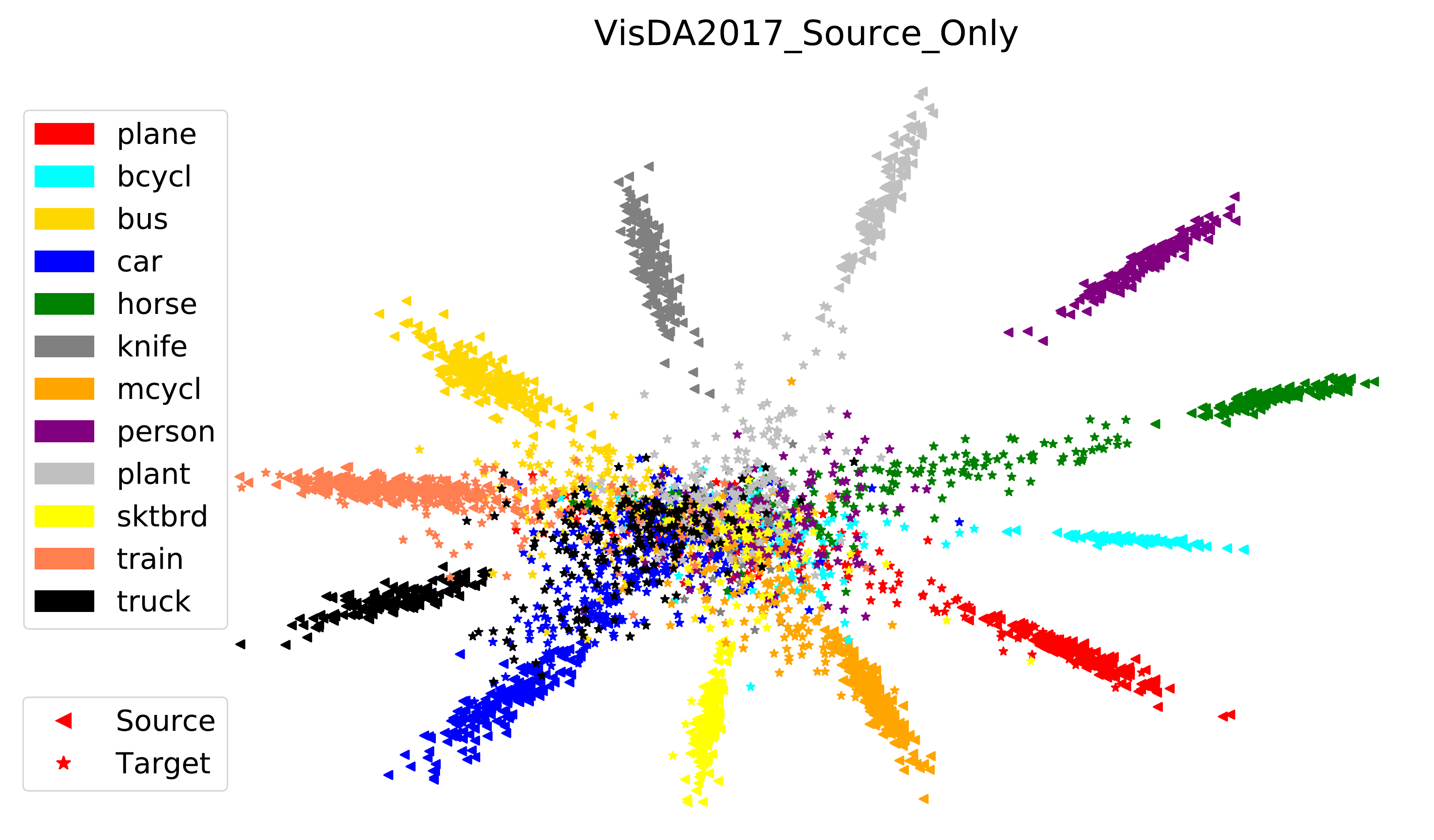}
    \includegraphics[width=0.48\textwidth]{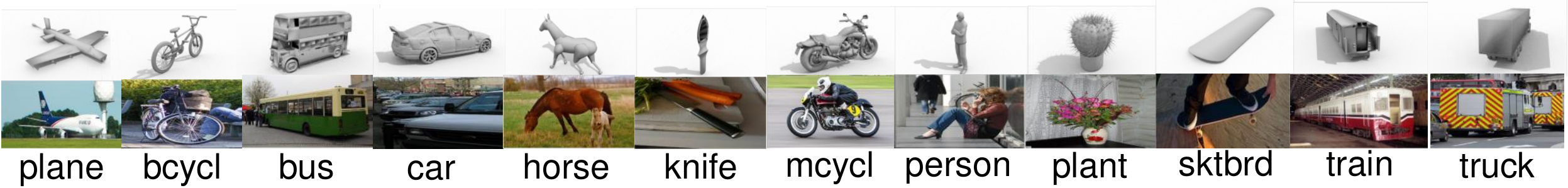}
\caption{\fontsize{9.5pt}{11.5pt}\selectfont Feature visualization of source and target samples on the Source Only model. This technique is widely used 
to characterize the feature embeddings under the softmax-related objectives~\cite{37_wen2016discriminative,38_liu2017sphereface, 64_zheng2018ring}. Specifically, we set the task-specific features to be two-dimensional and retrain the model. Unlike t-SNE~\cite{28_maaten2008visualizing} whose size of empty space does not account for the similarity between the two data points, this visualization map enables us to interpret the size of feature norms as well as inter-class and intra-class variances. As illustrated, target samples tend to collide in the small-norm (i.e., low-radius) regions which are vulnerable to slight angular variations of the decision boundaries and lead to erratic discrimination.} 
\label{fig:fea_visualization}
\end{figure}

\section{Introduction}
Deep neural networks, driven by numerous labeled samples, have made remarkable progress in a wide range of computer vision tasks. However, those models are very vulnerable to generalize into new application scenarios. Even a subtle deviation from the training regime can lead to a drastic degradation of the model~\cite{22_tzeng2017adversarial, 42_yosinski2014transferable}. Therefore, with the strong motivation to safely transfer knowledge from a label-rich source domain to an unlabeled target domain, Unsupervised Domain Adaptation (\textbf{UDA}) attempts to train a classifier using source samples that can generalize well to the target domain while mitigating the domain shift between the two underlying distributions. 

Under the guidance of the theoretical upper bound in~\cite{1_ben2010theory}, the key idea of most existing DA algorithms is to capture not only the task-discriminative but the domain-invariant representations by simultaneously minimizing the source error and some specific statistical discrepancy across the two domains, e.g., $\mathcal{H}$-divergence~\cite{1_ben2010theory, 4_ganin2014unsupervised, 22_tzeng2017adversarial}, 
$\mathcal{H}\Delta\mathcal{H}$-divergence~\cite{1_ben2010theory, 3_saito2017maximum}, Maximum Mean Discrepancy (\textbf{MMD})~\cite{5_borgwardt2006integrating, 18_tzeng2014deep, 2_long2015learning}, correlation distance~\cite{43_sun2016return, 44_sun2016deep} and etc. 

Adversarial domain adaptation~\cite{4_ganin2014unsupervised, 22_tzeng2017adversarial, 3_saito2017maximum, 45_hoffman2018cycada, 46_sankaranarayanan2018generate, 47_liu2017unsupervised, 49_bousmalis2016domain, 50_bousmalis2017unsupervised, 56_russo2018source}, which 
seeks to minimize an approximate domain discrepancy with an adversarial objective, has recently evolved into a dominant method in this field. To the best of our knowledge, RevGrad~\cite{4_ganin2014unsupervised} is the pioneer to empirically measure the $\mathcal{H}$-divergence by a parametric domain discriminator and adversarially align the features via reverse gradient backpropagation. ADDA~\cite{22_tzeng2017adversarial} instead facilitates the adversarial alignment with GAN-based objective in an asymmetric manner. MCD~\cite{3_saito2017maximum} places a min-max game between the feature generator and the two-branch classifiers to reduce the $\mathcal{H} \Delta \mathcal{H}$-divergence. On par with the feature-level alignment, generative pixel-level adaptation~\cite{45_hoffman2018cycada, 46_sankaranarayanan2018generate, 47_liu2017unsupervised, 50_bousmalis2017unsupervised, 56_russo2018source} utilizes Image-to-Image translation techniques to capture the low-level domain shifts.

While the notion of model degradation has been well recognized within the DA community~\cite{22_tzeng2017adversarial, 42_yosinski2014transferable}, little research work has been published to analyze the underlying cause of this phenomenon. Thus, existing statistic divergences may fail to precisely depict the domain shift and bridging such discrepancies may not guarantee the safe transfer across domains. For example, Shu \textit{et al}.~\cite{31_shu2018dirt} verify that bridging Jensen–Shannon divergence between the two domains does not imply high accuracy on the target task. In this paper, we take a step towards unveiling the nature of model degradation from a solid empirical observation, which is highlighted by Fig.~\ref{fig:fea_visualization}. This visualization map suggests that the excessively smaller norms of target-domain features with respect to that of the source domain account for their erratic discrimination. However, there remain two hypothetical interpretations from the current observation:

1) Misaligned-Feature-Norm Hypothesis: The domain shift between the source and target domains relies on their misaligned feature-norm expectations. Matching the mean feature norms of the two domains to an arbitrary shared 
scalar is supposed to yield similar transfer gains.

2) Smaller-Feature-Norm Hypothesis: The domain shift substantially relies on the excessively less-informative features with smaller norms for the target task. Despite non-rigorous alignment, adapting the target features far away from the small-norm regions can lead to safe transfer. 

With these points in mind, we introduce our parameter-free Adaptive Feature Norm (\textbf{AFN}) approach. First, we propose a simple yet effective statistic distance to characterize the mean-feature-norm discrepancy across domains. Second, we design the Hard AFN to bridge this domain gap by restricting the expected feature norms of the two domains to approximate a shared scalar. It suggests that norm-aligned features can bring effective transfer yet the results can be further improved with a larger scalar. To explore a more sufficient large feature norm in a stable way, we propose the Stepwise AFN to encourage a progressive feature-norm enlargement for each individual sample across domains. As stepwise AFN reveals, the key to achieving successful transfer is to properly lift the target samples towards the large-norm regions while the rigorous alignment is superfluous.

This innovative discovery inspires us to revisit what features are transferable. We recognize that those task-specific features with larger norms imply more informative transferability. Similar findings are explored in the field of model compression in terms of the smaller-norm-less-informative assumption~\cite{23_ye2018rethinking}, which suggests that parameters or features with smaller norms play a less informative role during the inference. Like the two sides of a coin, in contrast to the model compression that prunes unnecessary computational elements or paths, we place the larger-norm constraint upon the task-specific features to facilitate the more informative and transferable computation on the target domain.   

It is noteworthy that under the partial DA, the negative transfer is caused not only from the unrelated samples within the shared categories but also from the unrelated data from the source outlier categories. To this end, we propose meaningful protocols to evaluate the robustness w.r.t a specific algorithm to defense against these potential risks of negative transfer. With thorough evaluation, it reveals that our fairly novel feature-norm-adaptive manner is more robust to safely transfer knowledge from the source domain.


We summarize our contributions as follows:

i) We empirically unveil the nature of model degradation from a solid observation that the excessively smaller norms of the target-domain features with respect to that of the source domain account for their erratic discrimination.

ii) We propose a novel AFN approach for UDA by progressively adapting the feature norms of the two domains to a large range of scalars. Our approach is fairly simple yet effective and is translated into a few lines of code.

iii) We succeed in unifying the computation for both vanilla and partial DA and the feature-norm-adaptive manner is more robust to defense against the negative transfer. 

iv) Extensive experimental results have demonstrated the promise of our approach 
by exceeding state-of-the-arts 
across a wide range of visual DA benchmarks. 

\section{Related Work}
Domain adaptation~\cite{51_pan2010survey, 11_saenko2010adapting, 54_torralba2011unbiased} generalizes the learner across different domains by mitigating the domain shift problem. Supervised DA~\cite{53_tzeng2015simultaneous,
52_otiian_2017_ICCV, 55_motiian2017few} exploits a few labeled data in the target domain while unsupervised DA has no access to that. We focus on the latter scenario in our paper. 

Under the guidance of the theoretical upper bound proposed in~\cite{1_ben2010theory}, existing methods explore domain-invariant structures by minimizing some specific statistic distances between the two domains. For example, Maximum Mean Discrepancy (\textbf{MMD})~\cite{5_borgwardt2006integrating} based methods~\cite{2_long2015learning,13_long2017deep,18_tzeng2014deep} learn transferable features by minimizing the MMD of their kernel embeddings. Deep correlation alignment~\cite{44_sun2016deep} proposes to match the mean and covariance of the two distributions.~\cite{1_ben2010theory} introduces $\mathcal{H}$- and $\mathcal{H} \Delta \mathcal{H}$-divergence to characterize the domain discrepancy, which are further developed into matching the corresponding deep representations by~\cite{4_ganin2014unsupervised,22_tzeng2017adversarial} and~\cite{3_saito2017maximum} respectively. Regarding the methodology, kernel-based DA~\cite{15_gong2012geodesic, 57_gong2014learning, 2_long2015learning, 13_long2017deep} and adversarial DA~\cite{4_ganin2014unsupervised, 22_tzeng2017adversarial, 3_saito2017maximum, 45_hoffman2018cycada, 46_sankaranarayanan2018generate, 47_liu2017unsupervised, 50_bousmalis2017unsupervised, 56_russo2018source} are widely-used in the field.

Inspired by GANs~\cite{25_goodfellow2014generative}, adversarial DA involves a subnetwork as the domain classifier to discriminate features of different domains while the deep learner tries to generate the features that deceive the domain classifier. For example, RevGrad~\cite{4_ganin2014unsupervised} utilize a parametric subnetwork as the domain discriminator and adversarially align the features via reverse gradient backpropagation. ADDA~\cite{22_tzeng2017adversarial} instead facilitates the adversarial alignment with GAN-based objectives in an asymmetric manner. MCD~\cite{3_saito2017maximum} conducts a min-max game between the feature generator and the two-branch classifiers in order to reduce the 
$\mathcal{H} \Delta \mathcal{H}$-divergence. On par with the feature-level adversarial alignment, generative pixel-level adaptation~\cite{45_hoffman2018cycada, 46_sankaranarayanan2018generate, 47_liu2017unsupervised, 50_bousmalis2017unsupervised, 56_russo2018source} utilizes Image-to-Image translation techniques to capture the low-level domain shifts. 

In addition, other methods are proposed to learn target-specific structures. DRCN~\cite{19_ghifary2016deep} involves a reconstruction penalty on target samples.~\cite{33_saito2017asymmetric} utilizes tri-training to obtain target pseudo labels.~\cite{31_shu2018dirt} refines the target decision boundary based on the cluster assumption. 
iCAN~\cite{58_zhang2018collaborative} iteratively applies sample selection on pseudo-labeled target samples and retrains the network. 

Standard domain adaptation assumes that the two domains share the identical label space.~\cite{34_caopartial,9_cao2018partial} further open up the partial setting where source label space subsumes the target one,
However, it is not trivial to directly migrate the current models in the standard DA as they are prone to suffer from the negative transfer effect. PADA~\cite{9_cao2018partial} attempts to alleviate this issue by detecting and down-weighting samples belonging to the source outlier classes. 

\begin{figure}
    \setlength{\belowcaptionskip}{-0.36cm}
    \setlength{\abovecaptionskip}{0.05cm}
    \includegraphics[width=0.52\textwidth]{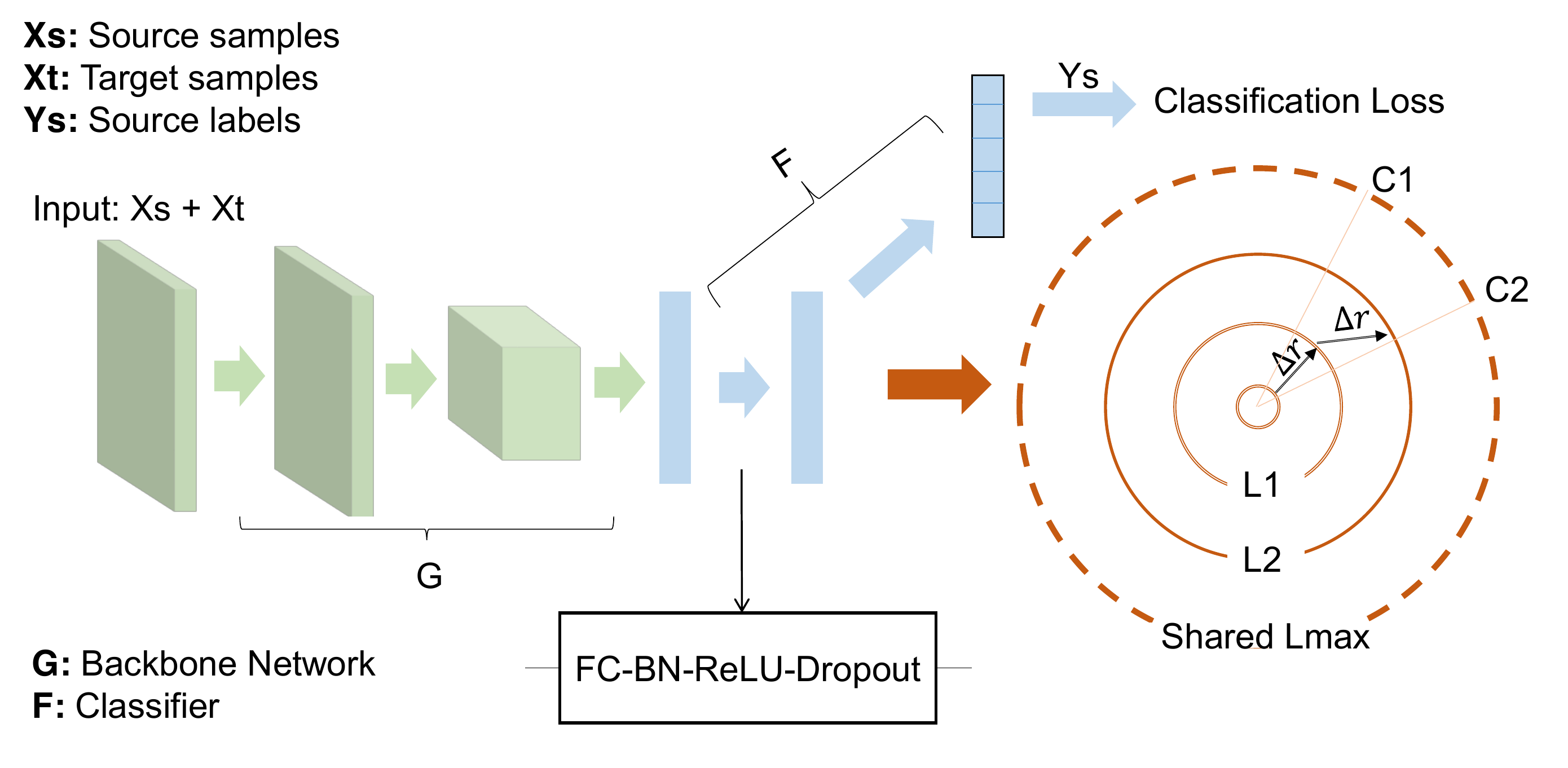}
    \caption{\fontsize{9.6pt}{11.5pt}\selectfont The overall framework of our proposed Adaptive Feature Norm approach. The backbone network $G$ denotes the general feature extraction module. $F$ is employed as the task-specific classifier with $l$ layers, each of which is organized in the FC-BN-ReLU-Dropout order. During each iteration, we apply the feature norm adaptation upon the task-specific features along with the source classification loss as our optimization objective. For the \textbf{Hard} variant of AFN, the mean feature norms of source and target samples are constrained to a shared scalar. For the \textbf{Stepwise} variant, we encourage a progressive feature-norm enlargement with respect to each individual example at the step size of $\Delta r$. To this end, far away from the small-norm regions after the adaptation, the target samples can be correctly classified without any supervision.}
\label{fig:framework}
\end{figure}

\section{Method}
\subsection{Preliminaries}
Given a source domain $\mathcal{D}_s = \{(x_i^s,y_i^s)\}_{i=1}^{n_s}$ of $n_s$ labeled samples associated with $|\mathcal{C}_s|$ categories and a target domain 
$\mathcal{D}_t = \{x_i^t\}_{i=1}^{n_t}$ of $n_t$ unlabeled samples associated with $|\mathcal{C}_t|$ categories. DA occurs when the underlying distributions corresponding to the source and target domains in the shared label space are 
\textit{different but similar}~\cite{24_ben2010impossibility} to make sense the transfer. Unsupervised DA considers the scenario that we have no access to any labeled target examples.

\textbf{Vanilla Setting} Under this setting, the source and target domains share the identical label space, i.e., $\mathcal{C}_s = \mathcal{C}_t$.

\textbf{Partial Setting} The source label space subsumes the target one, i.e., $\mathcal{C}_s \supset \mathcal{C}_t$. The source labeled data belonging to the outlier categories $\mathcal{C}_s \backslash \mathcal{C}_t$ are unrelated to the target task.

\subsection{\bm{$L_2$}-preserved Dropout}
In this part, we first prove that the standard Dropout operator is $L_1$-preserved. As our algorithm is computed based on the $L_2$ norms of the hidden features, we introduce the following $L_2$-preserved Dropout operation to meet our goal.  

Dropout is a widely-used regularization technique in deep neural networks~\cite{29_srivastava2014dropout, 26_li2018understanding}. Given a $d$-dimensional input vector \textbf{x}, in the training phase, we randomly zero the element $x_k,k = 1,2,$\ldots$,d$ with probability $\bm{p}$ by samples $a_k \sim P$ that are generated from the Bernoulli distribution:
\begin{equation}
    P(a_k)=
    \begin{cases}
    \quad p, & a_k=0\\
    1-p, & a_k=1
    \end{cases}
\end{equation}
To compute an identity function in the evaluation stage, the outputs are further scaled by a factor of $\frac{1}{1-p}$ and thus
\begin{equation}
    \setlength{\abovedisplayskip}{7pt}
    \setlength{\belowdisplayskip}{8pt}
    \begin{aligned}
        \hat{x}_{k}=a_k\frac{1}{1-p}x_k\,,
    \end{aligned}
\end{equation}
which implicitly preserves the $L_1$-norm in both of the training and evaluation phases since $x_k$ and $a_k$ are independent:
\begin{equation}
    \begin{aligned}
        \mathbb{E}[|\hat{x}_k|]=\mathbb{E}[|a_k\frac{1}{1-p}x_k|]= \frac{1}{1-p}\mathbb{E}[a_k]\mathbb{E}[|x_k|]=\mathbb{E}[|x_k|]\,.
    \end{aligned}
\end{equation}
However, as we are in pursuit of adaptive $L_2$ feature norm, we instead scale the output by a factor of $\frac{1}{\sqrt{1-p}}$ and obtain
\begin{equation}
    \setlength{\abovedisplayskip}{6pt}
    \setlength{\belowdisplayskip}{8pt}
    \begin{aligned}
    \hat{x}_k=a_k\frac{1}{\sqrt{1-p}}x_k\,,
    \end{aligned}
\end{equation}
which satisfies
\begin{equation}
    \small
    \begin{aligned}
        \mathbb{E}[|\hat{x}_k|^2]=\mathbb{E}[|a_k\frac{1}{\sqrt{1-p}}x_k|^2]=\frac{1}{1-p}\mathbb{E}[a_k^2]\mathbb{E}[|x_k|^2]=\mathbb{E}[|x_k|^2]\,.
    \end{aligned}
\end{equation}

\subsection{Framework}
As indicated in Fig.~\ref{fig:framework}, our framework consists of a backbone network $G$ and a classifier $F$. Existing findings reveal that deep features eventually transit from general to specific along the network and feature 
transferability significantly drops in higher layers~\cite{42_yosinski2014transferable}. In our case, $G$ is regarded as the general feature extraction module that inherits from the prevailing network architecture such as ResNet~\cite{16_he2016deep}. $F$ represents the task-specific classifier that has $l$ fully-connected layers. We denotes the first $l-1$ layers of the classifier as $F_f$, which results in the so-called bottleneck feature embeddings \textbf{f}~\cite{9_cao2018partial, 13_long2017deep}. Those features computed by $F_f$ depend greatly on the specific domain and are not safely transferable to a novel domain. Eventually, we calculate the class probabilities along the last layer $F_y$, which is followed by a softmax operator. We denote the parameters of $G$, $F_f$, $F_y$ with $\theta_g$, $\theta_f$ and $\theta_y$ respectively. Our intention is to explore an adaptive algorithm to compute the domain-transferable features 
\textbf{f} = $F_f(G(\cdot))$ using only source domain supervision. On the other side, as we are unifying the computation with respect to both vanilla and partial DA, it raises an interleaving challenge to defense against the negative transfer effect caused by the outlier categories in the source domain.
  
\subsection{Hard Adaptive Feature Norm}
Based on our Misaligned-Feature-Norm Hypothesis, we propose the Maximum Mean Feature Norm Discrepancy (\textbf{MMFND}) to characterize the mean-feature-norm distance between the two distributions and verify whether bridging this statistical domain gap can result in appreciable transfer gains. MMFND is defined by Eq.~(\ref{eq:mmfnd}), where the function class $\mathcal{H}$ is the combination of all the possible functions composited by the $L_2$-norm operator with the deep representation module, i.e.,
$ h(x) = (\left \| \cdot \right \|_2 \circ F_f \circ G)(x) $.
\begin{equation}
    \setlength{\abovedisplayskip}{5pt}
    \small
    \begin{aligned}
    \text{MMFND}[\mathcal{H},\mathcal{D}_s,\mathcal{D}_t] :=
    \sup_{h \in \mathcal{H}}(\frac{1}{n_s}\sum_{x_i\in \mathcal{D}_s}h(x_i)
    - \frac{1}{n_t}\sum_{x_i\in \mathcal{D}_t}h(x_i))\,.
    \end{aligned}
\label{eq:mmfnd}
\end{equation}

Intuitively, the functions class $\mathcal{H}$ are rich enough to contain substantial positive real valued functions on the input \textbf{x} and the upper bound will greatly deviate from zero if there is no restriction on the function type. In order to avoid this happening, we place a restrictive scalar $R$ to match the corresponding mean feature norms. By restricting both the mean feature norms of the two domains respectively converging towards the shared equilibrium $R$, the domain gap in terms of MMFND will vanish to zero. We implement this via the Hard Adaptive Feature Norm (\textbf{HAFN}) algorithm, which is illustrated by Eq.~(\ref{eq:bafn}).
\begin{equation}
    \setlength{\abovedisplayskip}{5pt}
    \small
    \begin{aligned}
    C_1(\theta_g,\theta_f,\theta_y)=\frac{1}{n_s}\sum_{(x_i,y_i) \in D_s}L_y(x_i, y_i) \\
    + \lambda(L_d(\frac{1}{n_s}\sum_{x_i \in D_s} h(x_i), R) + L_d(\frac{1}{n_t}\sum_{x_i \in D_t} h(x_i), R))\,.
    \end{aligned}
\label{eq:bafn}
\end{equation}

The optimization objective consists of two terms: the source classification loss $L_y$ in order to obtain the task-discriminative features by minimizing the softmax cross entropy on the source labeled samples, which is indicated by Eq.~(\ref{eq:cls}), where $p=\left[ p_1, \ldots, p_{|\mathcal{C}_s|} \right]$ is the softmax of the activations predicted by the classifier, i.e., $p=softmax(F(G(x)))$; the feature-norm penalty in order to obtain the domain-transferable features by minimizing the feature-norm discrepancy between the two domains, where $L_d(\cdot, \cdot)$ is taken as the $L_2$-distance and $\lambda$ is a hyper-parameter to trade off the two objectives.
\begin{equation}
    \setlength{\abovedisplayskip}{8pt}
    \setlength{\belowdisplayskip}{8pt}
    L_y(x_i^s,y_i^s;\theta_g,\theta_f,\theta_y)=-\sum_{k=1}^{|\mathcal{C}_s|}\mathds{1}_{[k=y_i^s]}\log p_k\,.
\label{eq:cls}
\end{equation}

Simple yet effective, MMFND appears to be a novel and superior statistical distance to characterize the cross-domain shift. And by bridging this feature-norm discrepancy with only source-domain supervision through executing HAFN, 
we can finally achieve the task-discriminative as well as domain-transferable features.

However, the preference setting of $R$ still remains unsettled. As the Misaligned-Feature-Norm Hypothesis suggests, matching feature-norm expectations of the two domains to an arbitrary shared positive real value is supposed to yield similar transfer gains. But this assertion is found not to be true by our empirical results. Specifically, although restricting the mean feature norms of the two domains to even a fairly small value (e.g., $R=1$, that is, feature normalization) has shown effective results, however, with $R$ gradually increases, the obtained models are still prone to achieve higher accuracies on the target task. To this end, it is natural to explore a sufficiently large $R$ and verify whether the rigorous alignment between the feature-norm expectations is necessary, which is revealed by our Smaller-Feature-Norm Hypothesis. In fact, it is unfortunate that HAFN fails to set an extremely large $R$ as the gradients generated by the feature-norm penalty may eventually lead to an explosion.

\subsection{Stepwise Adaptive Feature Norm}
To break the aforementioned bottleneck, we introduce an improved variant called Stepwise Adaptive Feature Norm (\textbf{SAFN}) in order to encourage the model to learn task-specific features with larger norms in a progressive manner, which is indicated by Eq.~(\ref{eq:safn}) as follows:
\begin{equation}
    \setlength{\abovedisplayskip}{5pt}
    \small
    \begin{aligned}
    C_2(\theta_g,\theta_f,\theta_y)=\frac{1}{n_s}\sum_{(x_i, y_i) \in D_s}L_y(x_i, y_i) \\
    + \frac{\lambda}{n_s+n_t}\sum_{x_i \in D_s \cup D_t} L_d(h(x_i;\theta_0)+\Delta r, h(x_i; \theta))\,,
    \end{aligned}
\label{eq:safn}
\end{equation}
where $\theta = \theta_g \cup \theta_f$. $\theta_0$ and $\theta$ represent the updated and updating model parameters in the last and current iterations respectively. $\Delta r$ denotes the positive residual scalar to control the feature-norm enlargement. During each iteration, the second penalty in SAFN encourages a feature-norm enlargement at the step size of $\Delta r$ with respect to individual examples, based on their feature norms calculated by the past model parameters in the last iteration. Instead of assigning a hard value, SAFN enables the optimization process more stable and fairly easy to trade off between the two objectives. To this end, executing SAFN can lead to higher accuracies on the target task by generating more informative features with larger norms. It is noteworthy that SAFN does not rigorously bridge the mean-feature-norm discrepancy, yet one can alternatively place a terminal $R$ to restrict the endless enlargement, which is indicated by Eq.~(\ref{eq:maxsafn}). However, our empirical results revealed that Eq.~(\ref{eq:maxsafn}) has slightly different result as to replace the second term in Eq.~(\ref{eq:safn}). As the Smaller-Feature-Norm hypothesis suggests, once we properly adapt the target samples towards the large-norm regions, the rigorous alignment becomes superfluous.
\begin{equation}
    \setlength{\abovedisplayskip}{5pt}
    \small
    \begin{aligned}
    L_d(\max(h(x_i;\theta_0)+\Delta r, R), h(x_i; \theta))\,.
    \end{aligned}
\label{eq:maxsafn}
\end{equation}



\begin{table*}[t]
    \addtolength{\tabcolsep}{-4pt}
    \centering
    \setlength{\abovecaptionskip}{0.0cm}   
    \caption{ Accuracy (\%) on \textit{Office-Home} under vanilla setting (ResNet-50)}
    \label{table:Office-Home-vanilla}
    \begin{small}
    \begin{tabular}{cccccccccccccc}
        \toprule[1.3pt]
        Method & Ar$\rightarrow$Cl & Ar$\rightarrow$Pr & Ar$\rightarrow$Rw & Cl$\rightarrow$Ar & Cl$\rightarrow$Pr & Cl$\rightarrow$Rw & Pr$\rightarrow$Ar & Pr$\rightarrow$Cl & Pr$\rightarrow$Rw & Rw$\rightarrow$Ar & Rw$\rightarrow$Cl & Rw$\rightarrow$Pr & Avg\\
        \hline
		ResNet~\cite{16_he2016deep} & 34.9 & 50.0 & 58.0 & 37.4 & 41.9 & 46.2 & 38.5 & 31.2 & 60.4 & 53.9 & 41.2 & 59.9 & 46.1\\
		DAN~\cite{2_long2015learning} & 43.6 & 57.0 & 67.9 & 45.8 & 56.5 & 60.4 & 44.0 & 43.6 & 67.7 & 63.1 & 51.5 & 74.3 & 56.3\\
		DANN~\cite{21_ganin2016domain} & 45.6 & 59.3 & 70.1 & 47.0 & 58.5 & 60.9 & 46.1 & 43.7 & 68.5 & 63.2 & 51.8 & 76.8 & 57.6\\
		JAN~\cite{13_long2017deep} & 45.9 & 61.2 & 68.9 & 50.4 & 59.7 & 61.0 & 45.8 & 43.4 & 70.3 & 63.9 & 52.4 & 76.8 & 58.3\\
        CDAN*~\cite{48_long2018conditional} & 49.0 & 69.3 & 74.5 & 54.4 & 66.0 & 68.4 & 55.6 & 48.3 & 75.9 & 68.4 & 55.4 & 80.5 & 63.8\\
        \toprule
        HAFN & 50.2$^{\pm0.2}$ & 70.1$^{\pm0.2}$ & \textbf{76.6}$^{\pm0.3}$ & 61.1$^{\pm0.4}$ & 68.0$^{\pm0.1}$ & 70.7$^{\pm0.2}$ & 59.5$^{\pm0.2}$ & 48.4$^{\pm0.3}$ & \textbf{77.3}$^{\pm0.2}$ & 69.4$^{\pm0.0}$ & 53.0$^{\pm0.6}$ & 80.2$^{\pm0.3}$ & 65.4\\
        SAFN & \textbf{52.0}$^{\pm0.1}$ & \textbf{71.7}$^{\pm0.6}$ & 76.3$^{\pm0.3}$ & \textbf{64.2}$^{\pm0.3}$ & \textbf{69.9}$^{\pm0.6}$ & \textbf{71.9}$^{\pm0.6}$ & \textbf{63.7}$^{\pm0.4}$ & \textbf{51.4}$^{\pm0.2}$ & 77.1$^{\pm0.0}$ & \textbf{70.9}$^{\pm0.4}$ & \textbf{57.1}$^{\pm0.1}$ & \textbf{81.5}$^{\pm0.0}$ & \textbf{67.3}\\
        \hline
        SAFN* & 54.4 & 73.3 & 77.9 & 65.2 & 71.5 & 73.2 & 63.6 & 52.6 & 78.2 & 72.3 & 58.0 & 82.1 & 68.5\\
        \toprule[1.3pt]
    \end{tabular}
    \end{small}
    \vspace{-1.0em}
\end{table*}
\begin{table*}[htbp]
    \addtolength{\tabcolsep}{-4.4pt}
    \centering
    \setlength{\abovecaptionskip}{0.0cm} 
    \caption{Accuracy (\%) on \textit{VisDA2017} under vanilla setting (ResNet-101)}
    \begin{small}
    \begin{tabular}{cccccccccccccc}
        \toprule[1.3pt]
        Method & plane & bcycl & bus & car & horse & knife & mcycl & person & plant & sktbrd & train & truck & Per-class \\
        \hline
		ResNet~\cite{16_he2016deep} & 55.1 & 53.3 & 61.9 & 59.1 & 80.6 & 17.9 & 79.7 & 31.2 & 81.0 & 26.5 & 73.5 & 8.5 & 52.4 \\
		DAN~\cite{2_long2015learning} & 87.1 & 63.0 & 76.5 & 42.0 & 90.3 & 42.9 & 85.9 & 53.1 & 49.7 & 36.3 & 85.8 & 20.7 & 61.1 \\
		DANN~\cite{21_ganin2016domain} & 81.9 & \textbf{77.7} & 82.8 & 44.3 & 81.2 & 29.5 & 65.1 & 28.6 & 51.9 & 54.6 & 82.8 & 7.8 & 57.4 \\
        MCD~\cite{3_saito2017maximum} & 87.0 & 60.9 & 83.7 & 64.0 & 88.9 & \textbf{79.6} & 84.7 & 76.9 & 88.6 & 40.3 & 83.0 & \textbf{25.8} & 71.9 \\
        \toprule
        HAFN & 92.7$^{\pm0.7}$ & 55.4$^{\pm4.1}$ & 82.4$^{\pm2.6}$ & \textbf{70.9}$^{\pm1.2}$ & 93.2$^{\pm0.9}$ & 71.2$^{\pm3.9}$ & 90.8$^{\pm0.5}$ & 78.2$^{\pm1.3}$ & 89.1$^{\pm0.7}$ & 50.2$^{\pm2.4}$ & 88.9$^{\pm0.8}$ & 24.5$^{\pm0.5}$ & 73.9 \\
        SAFN & \textbf{93.6}$^{\pm0.2}$ & 61.3$^{\pm4.0}$ & \textbf{84.1}$^{\pm0.5}$ & 70.6$^{\pm2.2}$ & \textbf{94.1}$^{\pm0.5}$ & 79.0$^{\pm4.1}$ & \textbf{91.8}$^{\pm0.5}$ & \textbf{79.6}$^{\pm1.3}$ & \textbf{89.9}$^{\pm0.7}$ & \textbf{55.6}$^{\pm3.4}$ & \textbf{89.0}$^{\pm0.3}$ & 24.4$^{\pm2.9}$ & \textbf{76.1} \\ 
        \toprule[1.3pt]
    \end{tabular}
    \end{small}
    \label{tab:acc_visda}
    \vspace{-1.3em}
\end{table*}
\subsection{Model Robustness Evaluation}\label{section:sec3.6}
Though the notion of \textit{negative transfer} has been well recognized within the DA community~\cite{51_pan2010survey}, its rigorous definition is still unclear~\cite{60_wang2018characterizing}. A widely accepted description of negative transfer~\cite{51_pan2010survey} is stated as “\textit{transferring knowledge from the source can have a negative impact on the target learner}”. While intuitive, how to evaluate it still remains open. Inspired by~\cite{60_wang2018characterizing}, we propose meaningful protocols to evaluate the robustness of a given algorithm especially under the more general partial setting. It is noteworthy that in this setting, the negative transfer is caused not only from the unrelated samples within the shared categories but also from the unrelated data from the source outlier classes. Let $A_{\mathcal{T}_{|\mathcal{C}_t|}}^{l\%}$, 
$A_{\mathcal{S}_{|\mathcal{C}_t|} \rightarrow \mathcal{T}_{|\mathcal{C}_t|}}$
and $A_{\mathcal{S}_{|\mathcal{C}_s|} \rightarrow \mathcal{T}_{|\mathcal{C}_t|}}$
denote the accuracies by using just $l\%$ target labeled data, transferring without and with source outlier classes w.r.t an identical algorithm. We define 
i) $A_{\mathcal{T}_{|\mathcal{C}_t|}}^{l\%}
- A_{\mathcal{S}_{|\mathcal{C}_t|} \rightarrow \mathcal{T}_{|\mathcal{C}_t|}}$
(Closed Negative Gap, \textbf{CNG}): the negative impact occurs if the algorithm cannot obtain more transfer gains over the negative influences from another domain than even just labeling a few (e.g., 1\%) target data, which is valueless when deployed in the wild.
ii) $A_{\mathcal{S}_{|\mathcal{C}_t|} \rightarrow \mathcal{T}_{|\mathcal{C}_t|}}
- A_{\mathcal{S}_{|\mathcal{C}_s|} \rightarrow \mathcal{T}_{|\mathcal{C}_t|}}$
(Outlier Negative Gap, \textbf{ONG}): especially measures the negative influences that are caused by the source unrelated categories.
iii) $A_{\mathcal{T}_{|\mathcal{C}_t|}}^{l\%}
- A_{\mathcal{S}_{|\mathcal{C}_s|} \rightarrow \mathcal{T}_{|\mathcal{C}_t|}}$
(Partial Negative Gap, \textbf{PNG}): reveals whether it is valuable for an algorithm to access and transfer from those available large domains with the potential risks of CNG and ONG. We say that the negative effect exceeds the positive gains once the gap value is positive and vice versa. The larger absolute value suggests more desperate negative influences or more encouraging positive gains.
\begin{table}[t]
    \addtolength{\tabcolsep}{-2.8pt}
    \centering
    \setlength{\abovecaptionskip}{0.0cm} 
    \caption{Accuracy (\%) on \textit{ImageCLEF-DA} in vanilla setting}
    \begin{small}
    \begin{tabular}{cccccccc}
        \toprule[1.3pt]
        Method & I$\rightarrow$P & P$\rightarrow$I & I$\rightarrow$C & C$\rightarrow$I & C$\rightarrow$P & P$\rightarrow$C & Avg\\
        \hline
		ResNet-50~\cite{16_he2016deep} & 74.8 & 83.9 & 91.5 & 78.0 & 65.5 & 91.2 & 80.7\\
        DAN~\cite{2_long2015learning} & 74.5 & 82.2 & 92.8 & 86.3 & 69.2 & 89.8 & 82.5\\
        DANN~\cite{21_ganin2016domain} & 75.0 & 86.0 & 96.2 & 87.0 & 74.3 & 91.5 & 85.0\\
        JAN~\cite{13_long2017deep} & 76.8 & 88.0 & 94.7 & 89.5 & 74.2 & 91.7 & 85.8\\
        CDAN*~\cite{48_long2018conditional} & 76.7 & 90.6 & \textbf{97.0} & 90.5 & 74.5 & 93.5 & 87.1\\
        \hline
        HAFN & 76.9 & 89.0 & 94.4 & 89.6 & 74.9 & 92.9 & 86.3 \\
        & $\pm$0.4 & $\pm$0.4 & $\pm$0.1 & $\pm$0.6 & $\pm$0.2 & $\pm$0.1 \\
        SAFN & 78.0 & 91.7 & 96.2 & 91.1 & 77.0 & 94.7 & 88.1 \\
        & $\pm$0.4 & $\pm$0.5 & $\pm$0.1 & $\pm$0.3 & $\pm$0.5 & $\pm$0.3 \\
        SAFN+ENT & \textbf{79.3} & \textbf{93.3} & 96.3 & \textbf{91.7} & \textbf{77.6} & \textbf{95.3} & \textbf{88.9} \\
        & $\pm$0.1 & $\pm$0.4 & $\pm$0.4 & $\pm$0.0 & $\pm$0.1 & $\pm$0.1 \\
        \hline
        SAFN+ENT* & 80.2 & 93.8 & 96.7 & 92.8 & 78.4 & 95.7 & 89.6 \\
        \toprule[1.3pt]
    \end{tabular}
    \label{tab:imageclef-da}
    \end{small}
    \vspace{-1em}
\end{table}
\begin{table}[t]
    \addtolength{\tabcolsep}{-4.6pt}
    \centering
    \setlength{\abovecaptionskip}{0.0cm}  
    \caption{Accuracy (\%) on \textit{Office-31} under vanilla setting}
    \begin{small}
    \begin{tabular}{cccccccc}
        \toprule[1.3pt]
        Method & A$\rightarrow$W & D$\rightarrow$W & W$\rightarrow$D & A$\rightarrow$D & D$\rightarrow$A & W$\rightarrow$A & Avg\\
        \hline
		ResNet-50~\cite{16_he2016deep} & 68.4 & 96.7 & 99.3 & 68.9 & 62.5 & 60.7 & 76.1\\
		DAN~\cite{2_long2015learning} & 80.5 & 97.1 & 99.6 & 78.6 & 63.6 & 62.8 & 80.4\\
        DANN~\cite{21_ganin2016domain} & 82.0 & 96.9 & 99.1 & 79.7 & 68.2 & 67.4 & 82.2\\
        ADDA~\cite{22_tzeng2017adversarial} & 86.2 & 96.2 & 98.4 & 77.8 & 69.5 & 68.9 & 82.9\\
        JAN~\cite{13_long2017deep} & 85.4 & 97.4 & 99.8 & 84.7 & 68.6 & 70.0 & 84.3\\
        GTA~\cite{46_sankaranarayanan2018generate} & 89.5 & 97.9 & 99.8 & 87.7 & 72.8 & \textbf{71.4} & 86.5\\
        CDAN*~\cite{48_long2018conditional} & \textbf{93.1} & 98.2 & \textbf{100.0} & 89.8 & 70.1 & 68.0 & 86.6\\
        \hline
        HAFN & 83.4 & 98.3 & 99.7 & 84.4 & 69.4 & 68.5 & 83.9\\
        & $\pm$0.7 & $\pm$0.1 & $\pm$0.1 & $\pm$0.7 & $\pm$0.5 & $\pm$0.3 \\
        SAFN & 88.8 & 98.4 & 99.8 & 87.7 & 69.8 & 69.7 & 85.7\\
        & $\pm$0.4 & $\pm$0.0 & $\pm$0.0 & $\pm$1.3 & $\pm$0.4 & $\pm$0.2 \\
        SAFN+ENT & 90.1 & \textbf{98.6} & 99.8 & \textbf{90.7} & \textbf{73.0} & 70.2 & \textbf{87.1}\\
        & $\pm$0.8 & $\pm$0.2 & $\pm$0.0 & $\pm$0.5 & $\pm$0.2 & $\pm$0.3 \\
        \hline
        SAFN+ENT* & 90.3 & 98.7 & 100.0 & 92.1 & 73.4 & 71.2 & 87.6\\
        \toprule[1.3pt]
    \end{tabular}
    \label{tab:acc_office}
    \end{small}
    \vspace{-1.3em}
\end{table}
\section{Experiment}
\subsection{Setup}

\textbf{VisDA2017}~\cite{6_peng2017visda} is the challenging large-scale benchmark that attempts to bridge the significant synthetic-to-real domain gap with over 280K images across 12 object categories. The source domain has 152,397 synthetic images generated by rendering from 3D models. The target domain has 55,388 real object images collected from \textit{Microsoft COCO}~\cite{7_lin2014microsoft}. Under the partial setting, we follow~\cite{9_cao2018partial} to choose (in alphabetic order) the first 6 categories as target categories and conduct the Synthetic-12 $\rightarrow$ Real-6 task.

\textbf{Office-Home}~\cite{10_venkateswara2017deep} is another challenging dataset that collects images of everyday objects to form four domains: Artistic images (\textbf{Ar}), Clipart images (\textbf{Cl}), Product images (\textbf{Pr}) and Real-World images (\textbf{Rw}). Each domain contains 65 object categories and they amount to around 15,500 images. Under the partial setting, we follow~\cite{9_cao2018partial} to choose (in alphabetic order) the first 25 categories as target categories. 

\textbf{Office-31}~\cite{11_saenko2010adapting} is a widely-used benchmark for visual DA. It contains 4,652 images of 31 office environment categories from three domains: \textit{Amazon} (\textbf{A}), \textit{DSLR} (\textbf{D}) and \textit{Webcam} (\textbf{W}), which correspond to online website, digital SLR camera and web camera images respectively. 

\textbf{ImageCLEF-DA} is built for ImageCLEF 2014 domain adaptation challenge\footnote[1]{\url{http://imageclef.org/2014/adaptation}} and consists of 12 common categories shared by three public datasets: \textit{Caltech-256} (\textbf{C}), \textit{ImageNet ILSVRC2012} (\textbf{I}) and \textit{Pascal VOC 2012} (\textbf{P}). There are 50 images in each category and 600 images in each domain. 

\begin{table*}[t]
    \addtolength{\tabcolsep}{-4pt}
    \centering
    \setlength{\abovecaptionskip}{0.0cm}   
    \caption{ Accuracy (\%) on \textit{Office-Home} under partial setting (ResNet-50)}
    \label{table:Office-Home-partial}
    \begin{small}
    \begin{tabular}{cccccccccccccc}
        \toprule[1.3pt]
        Method & Ar$\rightarrow$Cl & Ar$\rightarrow$Pr & Ar$\rightarrow$Rw & Cl$\rightarrow$Ar & Cl$\rightarrow$Pr & Cl$\rightarrow$Rw & Pr$\rightarrow$Ar & Pr$\rightarrow$Cl & Pr$\rightarrow$Rw & Rw$\rightarrow$Ar & Rw$\rightarrow$Cl & Rw$\rightarrow$Pr & Avg\\
        \hline
		ResNet~\cite{16_he2016deep} & 38.57 & 60.78 & 75.21 & 39.94 & 48.12 & 52.90 & 49.68 & 30.91 & 70.79 & 65.38 & 41.79 & 70.42 & 53.71\\
		DAN~\cite{2_long2015learning} & 44.36 & 61.79 & 74.49 & 41.78 & 45.21 & 54.11 & 46.92 & 38.14 & 68.42 & 64.37 & 45.37 & 68.85 & 54.48\\
		DANN~\cite{21_ganin2016domain} & 44.89 & 54.06 & 68.97 & 36.27 & 34.34 & 45.22 & 44.08 & 38.03 & 68.69 & 52.98 & 34.68 & 46.50 & 47.39\\
        PADA*~\cite{9_cao2018partial} & 51.95 & 67.00 & 78.74 & 52.16 & 53.78 & 59.03 & 52.61 & 43.22 & 78.79 & 73.73 & 56.60 & 77.09 & 62.06\\
        \hline
        \multirow{2}{23pt}{HAFN} & 53.35 & 72.66 & 80.84 & 64.16 & 65.34 & 71.07 & 66.08 & 51.64 & 78.26 & 72.45 & 55.28 & 79.02 & 67.51\\
        & $\pm$0.44 & $\pm$0.53 & $\pm$0.50 & $\pm$0.48 & $\pm$0.30 & $\pm$1.04 & $\pm$0.68 & $\pm$0.42 & $\pm$0.51 & $\pm$0.13 & $\pm$0.37 & $\pm$0.19 &\\
        \multirow{2}{23pt}{SAFN} & \textbf{58.93} & \textbf{76.25} & \textbf{81.42} & \textbf{70.43} & \textbf{72.97} & \textbf{77.78} & \textbf{72.36} & \textbf{55.34} & \textbf{80.40} & \textbf{75.81} & \textbf{60.42} & \textbf{79.92} & \textbf{71.83} \\
        & $\pm$0.50 & $\pm$0.33 & $\pm$0.27 & $\pm$0.46 & $\pm$1.39 & $\pm$0.52 & $\pm$0.31 & $\pm$0.46 & $\pm$0.78 & $\pm$0.37 & $\pm$0.83 & $\pm$0.20 &\\
        \hline
        SAFN* & 60.00 & 77.92 & 83.32 & 72.91 & 74.57 & 79.13 & 73.09 & 57.43 & 82.11 & 77.87 & 61.79 & 82.58 & 73.56\\
        \toprule[1.3pt]
    \end{tabular}
    \end{small}
    \vspace{-1.0em}
\end{table*}
\begin{table*}[t]
    \addtolength{\tabcolsep}{0pt}
    \centering
    \setlength{\abovecaptionskip}{0.0cm} 
    \caption{Evaluation on the Robustness}
    \begin{small}
    \begin{tabular}{ccccccccccccc}
        \toprule[1.3pt]
        \multirow{2}{*}{Method} & \multicolumn{3}{c}{Ar $\rightarrow$ Rw (5\%)} & \multicolumn{3}{c}{Cl $\rightarrow$ Rw (5\%)} 
        & \multicolumn{3}{c}{Pr $\rightarrow$ Rw (5\%)} & \multicolumn{3}{c}{VisDA2017 (1\%)} \\
        & CNG & ONG & PNG & CNG & ONG & PNG & CNG & ONG & PNG & CNG & ONG & PNG \\
        \hline
        DAN~\cite{2_long2015learning} & -14.6 & 13.8 & -0.8 & -4.9 & 24.5 & 19.6 & -12.0 & 17.3 & 5.3 & 18.7 & 25.0 & 43.7 \\
        DANN~\cite{21_ganin2016domain} & -17.2 & 21.9 & 4.7 & -10.9 & 39.4 & 28.5 & -14.5 & 19.5 & 5.0 & 13.6 & 26.7 & 40.3 \\
        JAN~\cite{13_long2017deep} & -15.3 & 13.1 & -2.2 & -7.1 & 19.1 & 12.0 & -13.4 & 12.2 & -1.2 & 20.4 & 24.5 & 44.9 \\
        PADA~\cite{9_cao2018partial} & -17.2 & 12.2 & -5.0 & -10.9 & 25.6 & 14.7 & -14.5 & 9.4 & -5.1 & 13.6 & 24.2 & 37.8 \\
        \hline
        SAFN & \textbf{-18.1} & \textbf{8.5} & \textbf{-9.6} & \textbf{-13.6} & \textbf{8.2} & \textbf{-5.4} & \textbf{-16.1} & \textbf{7.7} & \textbf{-8.4} & \textbf{5.0} & \textbf{15.6} & \textbf{20.6} \\
        \toprule[1.3pt]
    \end{tabular}
    \label{tab:robustness}
    \end{small}
    \vspace{-1.5em}
\end{table*}

\textbf{Implementation Details} We follow the standard protocol~\cite{2_long2015learning, 3_saito2017maximum, 4_ganin2014unsupervised, 21_ganin2016domain, 9_cao2018partial, 13_long2017deep, 22_tzeng2017adversarial} to utilize all labeled source data and unlabeled target data that belongs to their own label spaces. We implement our experiments on the widely-used 
\textbf{PyTorch}\footnote[2]{\url{https://pytorch.org/}} platform. For fair comparison, our backbone network is identical to the competitive approaches and is also fine-tuned from the ImageNet~\cite{35_deng2009imagenet} pre-trained model. We adopted a unified set of hyper-parameters throughout the \textit{Office-Home}, \textit{Office-31} and \textit{ImageCLEF-DA} datasets under both settings, where $\lambda = 0.05$, $R = 25$ in HAFN and $\Delta r = 1.0$ in SAFN. Since the synthetic domain on \textit{VisDA2017} is easy to converge, we applied a slightly smaller $\lambda$ and $\Delta r$ that equal to 0.01 and 0.3 respectively. We used mini-batch SGD optimizer with learning rate $1.0 \times 10^{-3}$ on all benchmarks. We used center-crop 
images for the reported results. For each transfer task, we reported the average accuracy over three random repeats. For fairer comparison with those methods~\cite{9_cao2018partial, 48_long2018conditional} which used ten-crop images at the evaluation phase with the best-performing models, we also included our corresponding results with the notion of \textbf{\{\textit{method}\}*} to benefit the future comparisons.

\begin{table}
    \addtolength{\tabcolsep}{0pt}
    \centering
    \setlength{\abovecaptionskip}{0.0cm} 
    \caption{Accuracy (\%) on \textit{VisDA2017} under partial setting}
    \label{table:partial-visda2017}
    \begin{small}
    \begin{tabular}{cc}
        \toprule[1.3pt]
        Method & Synthetic-12$\rightarrow$Real-6\\
        \hline
		ResNet-50~\cite{16_he2016deep} & 45.26\\
		DAN~\cite{2_long2015learning} & 47.60\\
        DANN~\cite{21_ganin2016domain} & 51.01\\
        PADA*~\cite{9_cao2018partial} & 53.53\\
        \hline
        HAFN & 65.06$\pm$0.90\\
        SAFN & \textbf{67.65}$\pm$0.51\\
        \hline
        SAFN* & 70.67\\
        \toprule[1.3pt]
    \end{tabular}
    \end{small}
    \vspace{-2.0em}
\end{table}
\subsection{Result Analysis}
Results on \textit{Office-Home}, \textit{VisDA2017}, \textit{ImageCLEF-DA} and \textit{Office-31} under the vanilla setting are reported in Table~\ref{table:Office-Home-vanilla},~\ref{tab:acc_visda},~\ref{tab:imageclef-da},~\ref{tab:acc_office} respectively. Results on \textit{Office-Home} and \textit{VisDA2017} under the partial setting are reported in Table~\ref{table:Office-Home-partial},~\ref{table:partial-visda2017}. Robustness evaluations in terms of CNG, ONG and PNG are shown in Table~\ref{tab:robustness}.
As illustrated, our methods significantly outperform the state-of-the-arts throughout all experiments, where SAFN is the top-performing variant.

Results on \textit{VisDA2017} reveal some interesting observations: i) Adversarial based models such as DANN may not effectively learn a diverse transformation across domains on this extremely large-scale transfer task and is prone to suffer from the risk of mode mismatch. However, our encouraging results prove the efficacy of AFN to work reasonably on this large-scale dataset and bridge the significant synthetic-to-real domain gap. ii) Note that existing methods usually mix and optimize multiple learning objectives, and it is not always easy to get an optimal solution. For example, MCD~\cite{3_saito2017maximum} incorporates another class balance objective to align target samples in a balanced way. Nevertheless, our method yields superior performance on most categories, revealing that it is robust to the unbalanced issue without any other auxiliary constraint. iii) Our model is parameter-free thus is more lightweight than the compared methods.

As indicated in Table~\ref{table:Office-Home-vanilla},~\ref{tab:imageclef-da} and~\ref{tab:acc_office}, our methods achieve new state-of-the-arts on these three benchmarks, and with larger rooms of improvement for those hard transfer tasks, e.g., D $\rightarrow$ A, A $\rightarrow$ D, Cl $\rightarrow$ Pr, Cl $\rightarrow$ Rw and etc, where the source and target domains are substantially different.

As illustrated in Table~\ref{table:Office-Home-partial} and~\ref{table:partial-visda2017}, our models obtain substantial improvements for partial DA, with 11.5\% gain on \textit{Office-Home} and 17.1\% gain on \textit{VisDA2017}. Plain domain-adversarial networks, e.g., DANN, seriously suffer from the mismatch from the source outlier classes and perform even worse than the Source Only variant. An intuitive solution, e.g., PADA~\cite{9_cao2018partial}, is to detect and down-weight the outlier categories during the domain alignment. However, without any heuristic reweighting mechanism, our feature-norm-adaptive manner exhibits stronger robustness against the unrelated data from the source domain. We testify this point via more thorough evaluation in Table~\ref{tab:robustness}. Besides, our method works stably and does not require to adjust different hyper-parameters for different subtasks within the same dataset as was done in PADA.

We carefully conduct robustness evaluation for the most challenging transfer tasks, e.g., Ar65 $\rightarrow$ Rw25, Synthetic-12 $\rightarrow$ Real-6 and etc. As described in Section~\ref{section:sec3.6}, the positive gap implies more negative impacts over the positive gains and vice versa. The target labeled ratio is 5\% and 1\% for the two benchmarks. Results in Table~\ref{tab:robustness} reveal some interesting observations: i) Throughout all evaluation metrics on all transfer tasks, we can either achieve the largest transfer gains or smallest negative influences. ii) All the methods, including ours, are inevitable to the positive ONG under the more challenging partial setting, while SAFN alleviates the outlier negative impact to the utmost extent. iii) For the Cl $\rightarrow$ Rw transfer task, the comparison methods all have positive PNG, suggesting that they are unable to obtain more transfer gains from the Cl65 domain than using only 5\% Rw25 labeled samples. However, we still derive encouraging result in this task. iv) It is noteworthy that on the most challenging \textit{VisDA2017} dataset with the significant synthetic-to-real gap, current approaches, including ours, all fail to distill more positive knowledge from the synthetic domain than just labeling 1\% real samples. It remains a big challenge for the future development of DA community.

\begin{figure*}[h]
    \vspace{0.2em}
    \begin{center}
    \subfigure[Sample Size]{
        \begin{minipage}[t]{0.236\linewidth}
        \centering
        \includegraphics[width=1\textwidth]{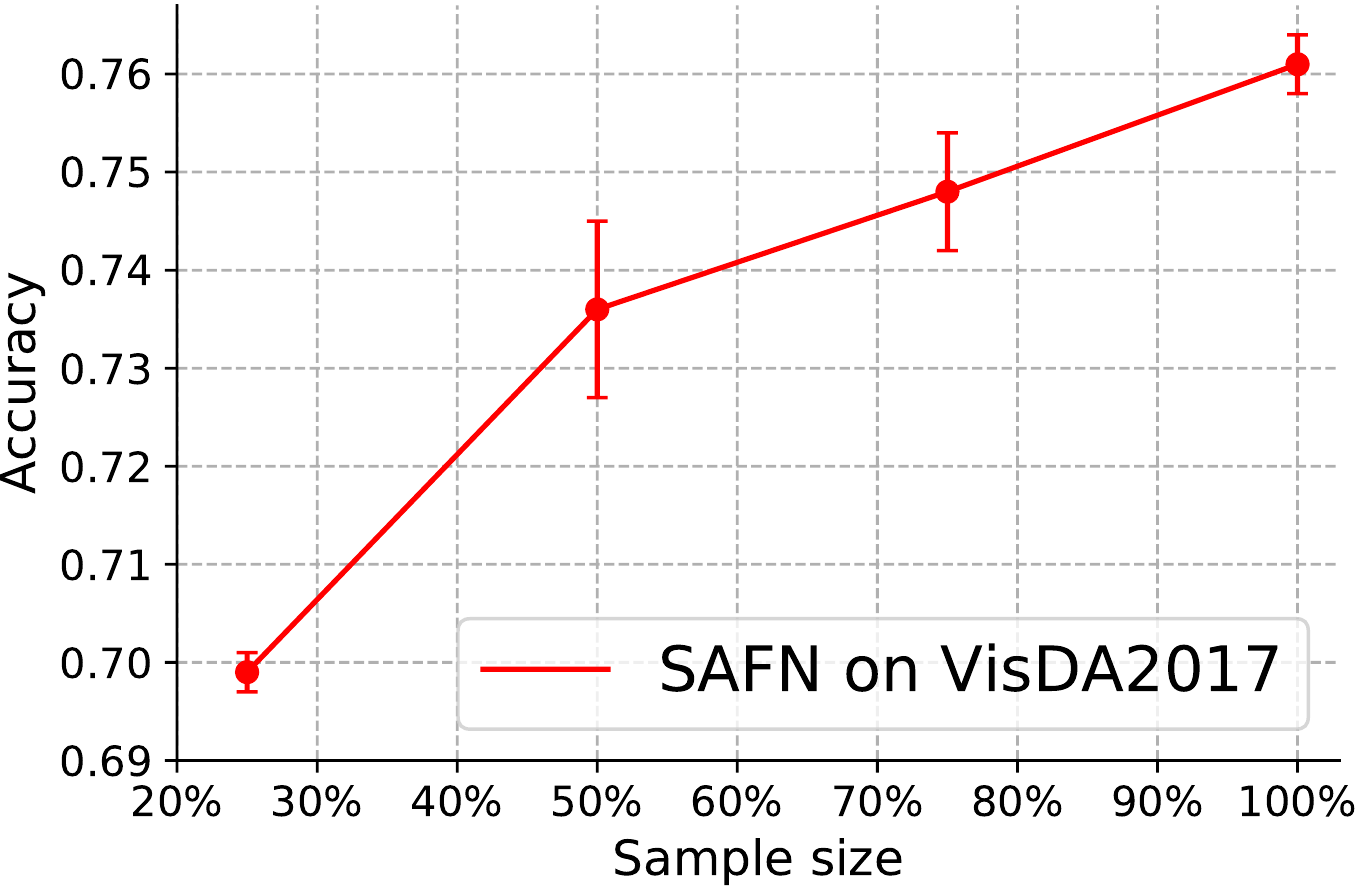}
        \label{fig:5a}
        \vspace{-1.2em}
        \end{minipage}
    }
    \subfigure[Sensitivity of R]{
        \begin{minipage}[t]{0.236\linewidth}
        \centering
        \includegraphics[width=1\textwidth]{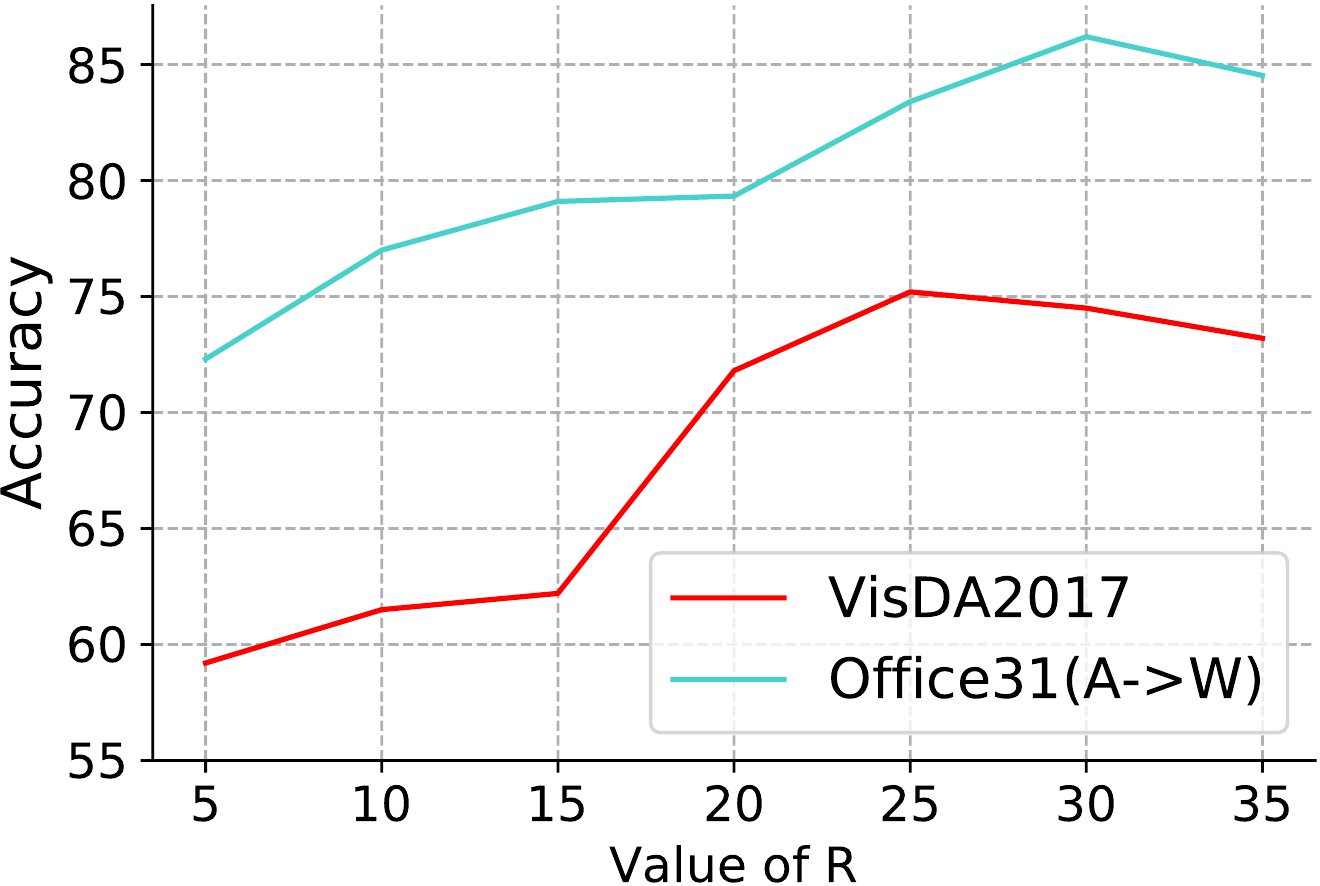}
        \label{fig:5b}
        \vspace{-1.2em}
        \end{minipage}
    }
    \subfigure[Sensitivity of $\Delta r$]{
        \begin{minipage}[t]{0.236\linewidth}
        \centering
        \includegraphics[width=1\textwidth]{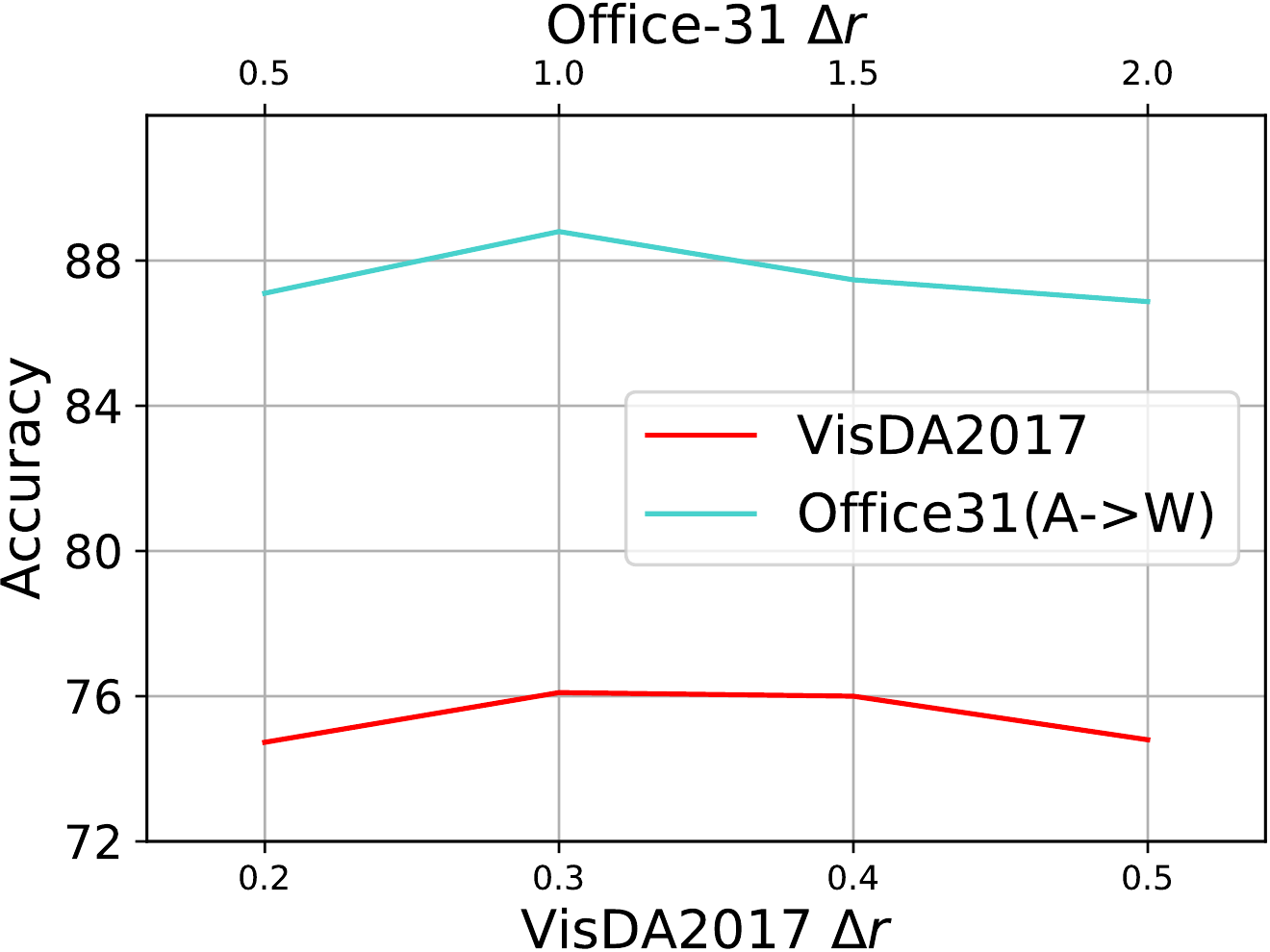}
        \label{fig:5c}
        \vspace{-1.2em}
        \end{minipage}
    }
    \subfigure[Embedding Size]{
        \begin{minipage}[t]{0.236\linewidth}
        \centering
        \includegraphics[width=1\textwidth]{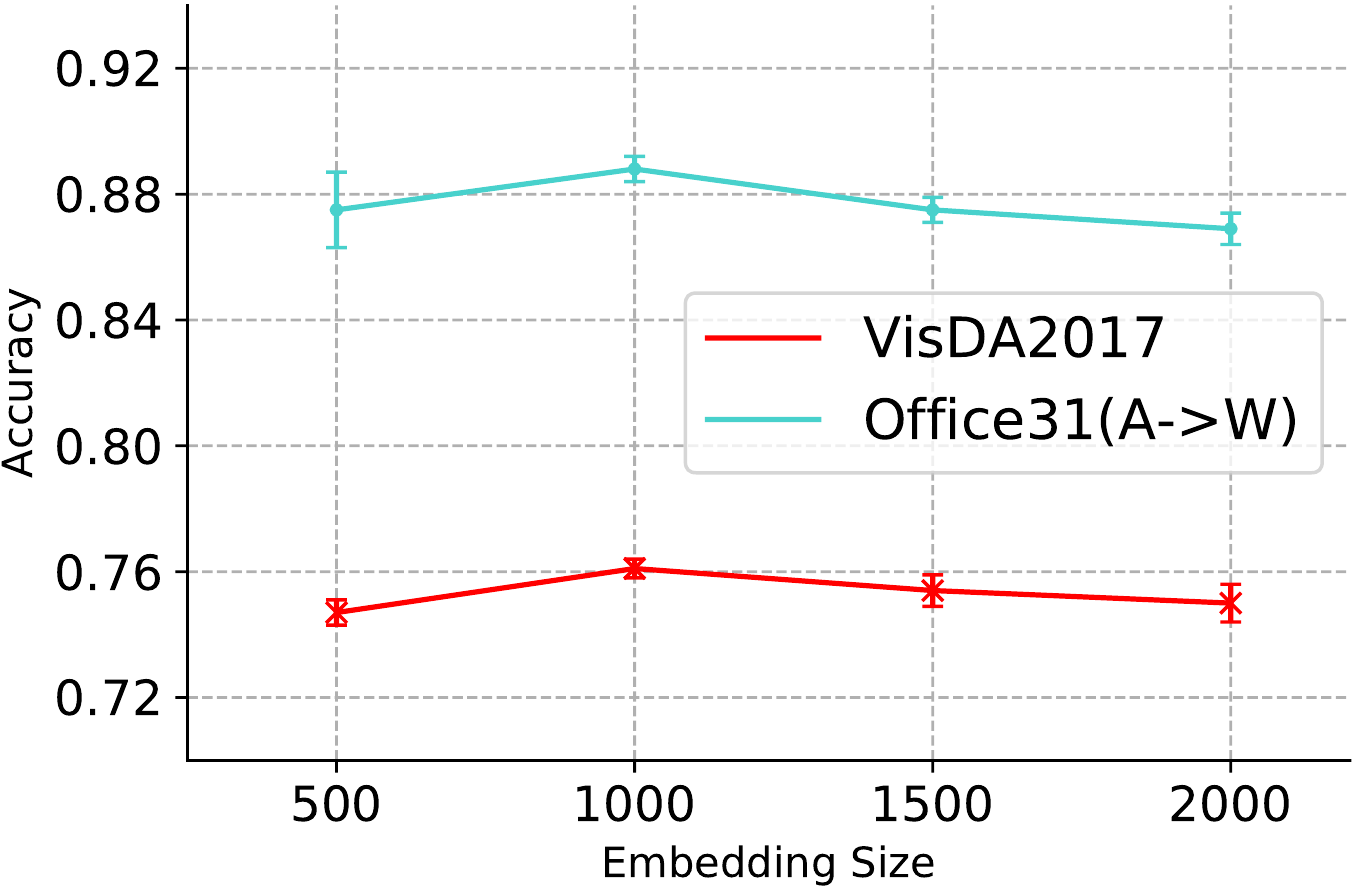}
        \label{fig:5d}
        \vspace{-1.2em}
        \end{minipage}
    }
    \end{center}
    \setlength{\abovecaptionskip}{-0.4cm}   
    \setlength{\belowcaptionskip}{-0.44cm}   
    \caption{\fontsize{9.4pt}{11.4pt}\selectfont Analysis of (a) varying unlabeled target sample size; (b)(c) parameter sensitivity of $R$ and $\Delta r$; (d) varying embedding size.}
    \label{fig:visualization}
\end{figure*}
\begin{figure}[h]
    \vspace{0.2em}
    \begin{center}
    \subfigure[Source Only]{
        \begin{minipage}[t]{0.474\linewidth}
        \centering 
        \includegraphics[width=1\textwidth]{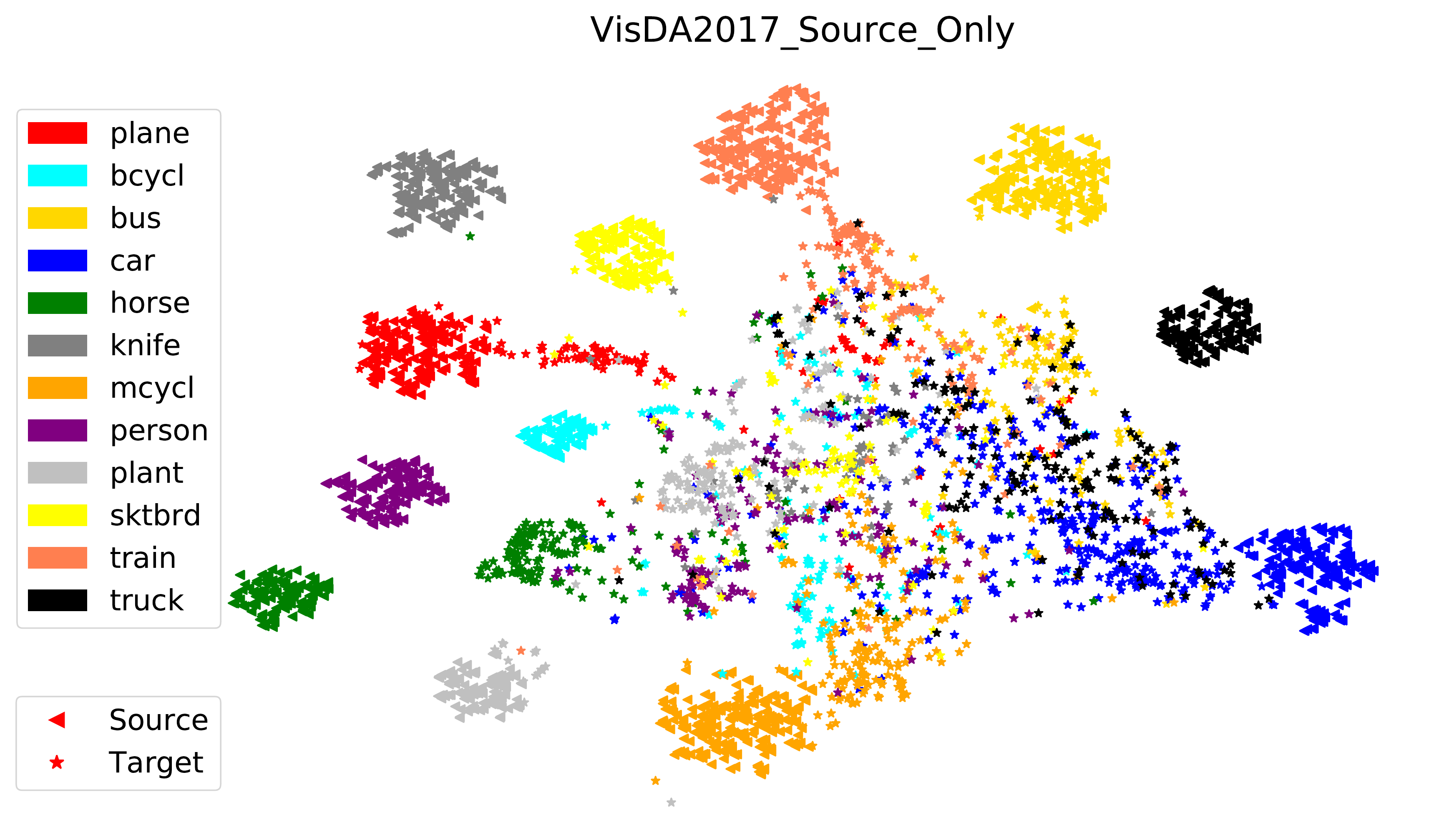}
        \label{fig:4a}
        \vspace{-1.2em}
        \end{minipage}
    }
    \subfigure[SAFN]{
        \begin{minipage}[t]{0.474\linewidth}
        \centering
        \includegraphics[width=1\textwidth]{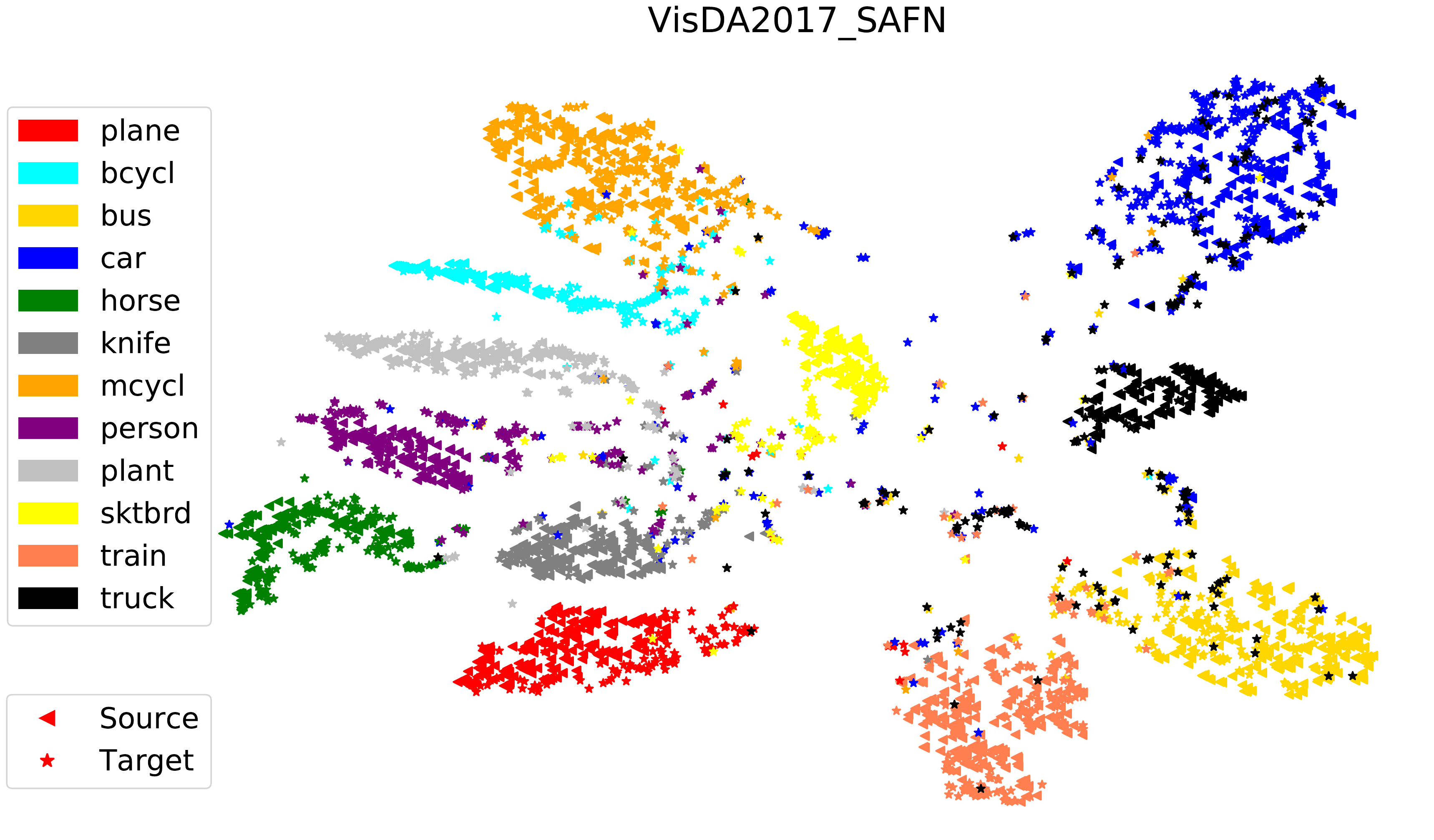}
        \label{fig:4b}
        \vspace{-1.2em}
        \end{minipage}
    }
    \end{center}
    \setlength{\abovecaptionskip}{-0.4cm}   
    \setlength{\belowcaptionskip}{-0.5cm}   
    \caption{\fontsize{9.4pt}{11.4pt}\selectfont 
    (a) and (b) correspond to the t-SNE embedding visualization of the Source Only and SAFN models on \textit{VisDA2017}. The triangle and star markers denote the source and target samples respectively. Different colors indicate different categories.}
    \label{fig:visualization}
\end{figure}
\section{Ablation Study}
\textbf{Feature Visualization:} Although testifying the efficacy of a DA 
algorithm via t-SNE~\cite{28_maaten2008visualizing} embeddings is considered 
over-interpreted,\footnote[3]{\url{https://interpretablevision.github.io/}}
we still follow the \textit{de facto} practice to provide the intuitive 
understanding. We randomly select 2000 samples across 12 categories from the 
source and target domains on \textit{VisDA2017} and visualize 
their task-specific features by t-SNE. As shown in Fig.~\ref{fig:4a}, the 
ResNet features of target samples collide into a mess because of extremely 
large synthetic-to-real domain gap. After adaptation, as illustrated in 
Fig.~\ref{fig:4b}, our method succeeded in separating target domain samples 
and better aligning them to the corresponding source domain clusters.

\textbf{Sample Size of Target Domain:} In this part, we empirically 
demonstrate that our approach is scalable and data-driven with respect to the increase of~\textbf{unlabeled} target samples, which exposes the 
appealing capacity in practice. It is not necessarily intuitive for 
adversarial learning based methods to optimize and obtain a diverse 
transformation upon large volumes of unlabeled target samples. Specifically, 
we shuffle the target domain on \textit{VisDA2017} and sequentially access 
the top 25\%, 50\%, 75\% and 100\% of the dataset. We train and evaluate 
our approach on these four subsets. As illustrated in Fig.~\ref{fig:5a}, 
with the sample size gradually increases, the classification accuracy of the 
corresponding target domain grows accordingly. It shows that the more unlabeled target data are involved in the feature norm adaptation, the more transferable classifier can be obtained.

\textbf{Complementary with Other Methods:} In this part, we demonstrate that 
our approach can be used in combination with other DA techniques. Because of 
limited space, we particularly exploit ENTropy minimization 
(\textbf{ENT})~\cite{30_grandvalet2005semi}, a low-density separation 
technique, for demonstration. \textbf{ENT} is widely applied in DA 
community~\cite{17_long2016unsupervised,31_shu2018dirt,32_long2018transferable} 
to encourage the decision boundary to pass through the target low-density 
regions by minimizing the conditional entropy of target samples. We conduct 
this case study on \textit{ImageCLEF-DA} and \textit{Office-31} datasets and 
report the accuracies in Table~\ref{tab:imageclef-da} and~\ref{tab:acc_office} 
respectively. As indicated, with \textbf{ENT} 
to fit the target-specific structure, we further boost the recognition 
performance by 0.8\% and 1.4\% 
on these two datasets.

\textbf{Sensitivity of $R$ and $\Delta r$:} We conduct case studies to 
investigate the sensitivity of parameter $R$ in HAFN and parameter 
$\Delta r$ in SAFN. We select \textit{VisDA2017} and task A$\rightarrow$W 
for demonstration. The results are shown in Fig.~\ref{fig:5b} and 
Fig.~\ref{fig:5c}, by varying $R\in$\{5, 10, 15, 20, 25, 30, 35\} on both 
datasets, $\Delta r\in$ \{0.2, 0.3, 0.4, 0.5\} on \textit{VisDA2017} and 
$\Delta r\in$ \{0.5, 1.0, 1.5, 2.0\} on task A$\rightarrow$W. 
For parameter $R$, the accuracy first gradually increases with larger values 
of $R$ and then begins to decrease as the feature-norm penalty in HAFN may explode.
As shown in Fig.~\ref{fig:5c}, the accuracies stay almost the same 
as parameter $\Delta r$ varies, revealing that SAFN works reasonably stable
on these two tasks.  

\textbf{Sensitivity of Embedding Size:} We investigate the sensitivity of 
embedding size of the task-specific features as it plays a significant role 
in norm computation. We conduct this case study on both 
\textit{VisDA2017} and A$\rightarrow$W transfer tasks. 
We report the average accuracy over three random repeats for those embedding
sizes varying in \{500, 1000, 1500, 2000\}. As illustrated in 
Fig.~\ref{fig:5d}, the accuracy stays almost the same and achieves slightly 
higher when the embedding size is set to 1000, indicating that our approach is robust to a wide range of feature space dimensions.

\section{Conclusion}
We have presented an innovative discovery for UDA, revealing that the model degradation on the target domain mainly stems from its much smaller feature norms with respect to that of the source domain. To this end, we demonstrated that progressively adapting the feature norms of the two domains to a large range of values can result in significant transfer gains, implying that those task-specific features with larger norms are more transferable. Our method is parameter-free, easy to implement and performs stably. In addition, we successfully unify the computation of both standard and partial DA, and thorough evaluations revealed that our feature-norm-adaptive manner is more robust against the negative transfer. Extensive experimental results have validated the virtue of our proposed approach.
\section*{Acknowledgements}
This work was supported in part by the State Key Development Program under Grant 2016YFB1001004, in part by the National Natural Science Foundation of China under Grant No.U1811463 and No.61702565, in part by  Guangdong “Climbing Program” Special Funds (pdjhb0009). This work was also sponsored by SenseTime Research Fund.

{\small
\bibliographystyle{ieee}
\bibliography{egbib}
}

\end{document}